# The Enduring Dominance of Deep Neural Networks: A Critical Analysis of the Fundamental Limitations of Quantum Machine Learning and Spiking Neural Networks

Takehiro Ishikawa[1]*

[1]College of Computing, Georgia Institute of Technology, Atlanta, GA, USA

*Corresponding author: tishikawa8@gatech.edu

Recent advancements in quantum machine learning (QML) and spiking neural networks (SNNs) have generated considerable excitement, promising exponential speedups and brain-like energy efficiency to revolutionize artificial intelligence (AI). However, this paper critically examines their fundamental limitations, arguing that they are unlikely to displace deep neural networks (DNNs) in the near term. QML struggles with adapting backpropagation due to unitary constraints, measurement-induced state collapse, barren plateaus, and high measurement overheads, exacerbated by the limitations of current noisy intermediate-scale quantum hardware, overfitting risks due to underdeveloped regularization techniques, and a fundamental misalignment with machine learning's generalization. SNNs face restricted representational bandwidth, struggling with long-range dependencies and semantic encoding in language tasks due to their discrete, spike-based processing, unlike the attention mechanisms of Transformers. Furthermore, the goal of faithfully emulating the brain might impose inherent inefficiencies like cognitive biases, limited working memory, and slow learning speeds. Even their touted energy-efficient advantages are overstated; optimized DNNs with quantization can outperform SNNs in energy costs under realistic conditions. Finally, SNN training incurs high computational overhead from temporal unfolding. In contrast, DNNs leverage efficient backpropagation, robust regularization, and innovations in large reasoning models that shift scaling to inference-time compute, enabling self-improvement via reinforcement learning and search algorithms like Monte Carlo tree search while mitigating data scarcity. This superiority is evidenced by recent models such as xAI's Grok-4 Heavy, which advances state-of-the-art performance, and gpt-oss-120b, which surpasses or approaches the performance of leading industry models like OpenAI's o3-mini and o4-mini despite its modest 120-billion-parameter size deployable on a single 80GB GPU. Furthermore, specialized application-specific integrated circuits, such as the Cerebras Wafer-Scale Engine, the Groq Language Processing Unit, and the Etched Sohu, amplify these efficiency gains. Ultimately, QML and SNNs may serve niche hybrid roles, but DNNs remain the dominant, practical paradigm for AI advancement.

## 1. Introduction

Recent years have witnessed significant enthusiasm surrounding emerging computational paradigms such as quantum computing and spiking neural networks (SNNs). This surge in interest is particularly evident in Quantum Machine Learning (QML), where hype suggests that integrating quantum computing with Artificial Intelligence (AI) could revolutionize AI by providing exponential speedups for complex tasks such as optimization, big data analysis, and machine learning training, potentially solving problems intractable on classical computers[1]. Similarly, SNNs have attracted attention due to their potential to mimic the energy efficiency of the human brain—which operates on a mere 20 watts[2] —especially amid growing concerns over the excessive power consumption of current AI systems, where inference for large language models (LLMs) like GPT-4o can equate to the annual electricity usage of tens of thousands of households at scale[3]. This has fueled speculation that more faithful replication of neural processes could address these sustainability challenges and lead to the next major breakthrough in artificial intelligence[2].

Despite the growing hype, the realities remain far more challenging. This paper critically examines the core limitations of QML and SNNs, while underscoring the practical strengths of Deep Neural Networks (DNNs)—the prevailing paradigm in the AI industry[4]—to argue that QML and SNNs are unlikely to displace DNNs anytime soon.

QML grapples with challenges such as the difficulty in adapting backpropagation—a key algorithm that propagates errors backward through each layer of a neural network, efficiently computing gradients for all parameters and enabling the training of large-scale networks, which has underpinned modern AI breakthroughs[5]. This difficulty arises due to hurdles like implementing nonlinear operations[6], measurement-induced state collapse[7-10], barren plateaus[11], and the steep measurement overhead in parameter-shift rules[7]. Moreover, these issues are further compounded by overfitting risks arising from the underdevelopment of regularization mechanisms[12], the nascent state of quantum hardware[10] and its inherently different nature compared to the generalization capabilities of classical machine learning[13].

Meanwhile, SNNs are hindered by restricted representational bandwidth and challenges in handling long-range dependencies for language tasks[14-15]. Furthermore, faithfully emulating the brain's evolutionarily constrained mechanisms introduces redundant and inefficient processes, such as cognitive biases, limited working memory, and slow learning speeds, which hinder rapid scaling and high performance compared to silicon-based AI[16-19]. Additionally, claims of superior energy efficiency are often overstated, as optimized DNNs using quantization can outperform SNNs in practice under realistic conditions[20-21], and SNN training incurs high computational and memory overhead due to temporal unfolding across time steps, often necessitating conversion from pre-trained DNNs[21-23].

In contrast, the dominance of DNNs stems from efficient training via backpropagation, a method that enables the scaling of massive models like Transformers by computing all parameter gradients in a single pass[5]. This efficiency is complemented by robust regularization techniques, such as L1/L2 penalties and dropout, which effectively prevent overfitting and improve generalization[12,17-18].

Looking ahead, DNNs' future prospects are bolstered by innovations in Large Reasoning Models (LRMs). These models pioneer a shift from a data-centric to a compute-centric scaling paradigm by dedicating more resources to inference-time reasoning and self-generated data via reinforcement learning and Monte Carlo Tree Search (MCTS) [24-25]. The recent release of xAI's Grok-4 Heavy exemplifies this, setting new industry standards with state-of-the-art results: 88.9% on GPQA, 100% on AIME 2025 (with tools), and 44.4% on Humanity's Last Exam (with tools) [26].

This strategy mitigates data scarcity and reduces pretraining costs. It enables smaller models to outperform larger ones within the same compute budget by employing techniques like Mixture-of-Experts (MoE), quantization, and knowledge distillation[27-31]. The success of this approach is evident in models like gpt-oss-120b. Despite its relatively modest 120-billion-parameter size and ability to run on a single 80GB GPU, it rivals leading industry models on key benchmarks. For instance, it achieves 97.9% on AIME 2025 (with tools), 80.1% on GPQA Diamond (without tools), and 90.0% on MMLU, consistently outperforming OpenAI's o3-mini and approaching the capabilities of o4-mini[31].

Complementing this, the transition to specialized Application-Specific Integrated Circuits (ASICs)—such as Cerebras WSE, Groq's LPU, and Etched's Sohu—promises dramatic improvements in inference efficiency. For instance, Cerebras WSE offers 10–20× latency reductions and 2.5× better energy efficiency compared to GPUs[32]. According to self-reported benchmarks, Groq's LPU achieves token generation rates exceeding 300 per second on large models like Llama-2 70B, providing inference that is 10 times faster and more energy-efficient than traditional GPU setups[33]. This advantage is confirmed by independent research, which found up to 20× lower latency than NVIDIA A100 GPUs[34]. Finally, based on pre-commercial claims ahead of its 2025 release, a single 8xSohu server from Etched is projected to serve over 500,000 Llama 70B tokens per second. This performance would match 160 H100 GPUs while being an order of magnitude faster and cheaper than NVIDIA's B200[35].

The novelty of this paper lies in contrasting the systematically organized challenges of QML and SNNs with DNNs' mature ecosystem and in highlighting their advantages from a practical perspective—robustly supported by drawing on numerous papers from 2025 and cutting-edge industry trends, particularly in LRMs and ASICs.

**2. Challenges of QML**

**2.1. Difficulty in Applying Backpropagation to QML**

*2.1.1. Unitary Operations*

Aside from measurement and encoding steps, quantum circuits inherently allow only unitary transformations. In contrast, backpropagation in classical neural networks relies on flexibly customizable, nonlinear activation functions. Because no straightforward quantum equivalent exists for these nonlinear operations, applying conventional backpropagation methods directly to quantum circuits becomes challenging[6].

*2.1.2. State Collapse due to Measurement*

A core challenge in quantum computing is that any measurement collapses the quantum state, making it impossible to store and retrieve intermediate results in the same way as in classical backpropagation. In classical backpropagation, a key algorithm for training neural networks, intermediate results—such as the activation values from each layer during the forward pass—are temporarily stored in memory and later referenced during the backward pass to efficiently calculate gradients and update model parameters [5]. However, in quantum systems, accessing these intermediate quantum states via measurement causes the fragile superposition to collapse irreversibly [7]. One might consider cloning the quantum state before measurement, but the no-cloning theorem prohibits creating perfect copies of unknown quantum states[8]. Given these constraints, Generative Quantum Machine Learning (GQML), such as Quantum Generative Adversarial Networks (QGANs), provides an alternative path forward. Instead of attempting to replicate quantum states—which is explicitly forbidden by the no-cloning theorem—GQML aims to learn target distributions and approximately generate new quantum states sampled from them [9]. Still, the limitations of current Noisy Intermediate-Scale Quantum (NISQ) devices, such as qubit decoherence and the complexity of error correction, severely hinder long-term storage and manipulation of quantum information [10]. Together, these fundamental principles and practical engineering challenges form substantial barriers to applying classical backpropagation directly within quantum circuits.

*2.1.3. Barren Plateau*

The barren plateau problem refers to a phenomenon often encountered in variational quantum circuits, where the gradient of the cost function with respect to circuit parameters becomes extremely small— often vanishing—making training prohibitively difficult. This effect is especially pronounced as the number of qubits and the circuit depth increase. In the initial stages, when data or features are encoded into quantum states, some structure or directional bias may indeed be present. However, once the circuit parameters are randomly initialized and multiple layers of parameterized unitary gates act on the states, the measurement statistics become highly "scrambled." Because unitary transformations preserve norms but randomize phases and amplitudes, the overall distribution of measurement outcomes tends to appear uniform on average. Consequently, the gradient of the cost function with respect to each parameter collapses toward zero, giving rise to a nearly flat optimization landscape. In such scenarios, even small or local updates to the parameters fail to reduce the cost function in any meaningful way, and training effectively stalls [11].

*2.1.4. Measurement Overhead in Parameter-shift Rules*

In classical backpropagation, a single forward pass and backward pass compute the gradients with respect to all parameters simultaneously, enabling efficient gradient-based optimization[5]. In contrast, when using the parameter shift rule for quantum circuits, the gradient for each parameter is obtained by running the circuit twice: once with a slightly increased parameter value and once with a slightly decreased value. By measuring the outputs for these two shifted configurations, one can estimate the gradient for that specific parameter. However, this procedure must be repeated independently for every parameter. Consequently, if a circuit has N parameters, it requires 2N separate executions and measurements to determine all gradients. The resulting increase in measurement operations and circuit runs significantly extends the time and resources needed as the number of parameters grows, surpassing what is typically required in classical approaches[7].

## 2.2. Overfitting Risks and Underdeveloped Regularization Mechanisms

While QML frameworks offer a high representational capacity by leveraging the vast dimensionality of Hilbert space, this capacity alone does not guarantee strong generalization performance. In the same way that Transformer models excel not merely due to their representational power but also because of large training datasets and effective regularization strategies (e.g., L1/L2 penalties, dropout), quantum models likewise require mechanisms to prevent overfitting and enhance their generalization capabilities, yet these mechanisms remain underdeveloped [12].

## *2.3.* Limitations of Small-Scale Benchmarks

A significant challenge in QML arises from the early stage of quantum technology. Currently, quantum devices can accommodate only a small number of qubits and allow for limited-depth circuits, constraining experiments to smaller-scale benchmarks. However, success on trivial examples does not necessarily translate to more complex tasks; hence, small-scale quantum benchmarks may offer limited insight into how such methods will perform on more challenging problems[13].

## 2.4. Fundamental Misalignment with Machine Learning's Generalization

Quantum computing excels at efficiently solving highly structured, well-defined problems that often exhibit periodic structures exploitable by quantum interference, such as integer factorization via Shor's algorithm. Other applications include unstructured database searches via Grover's algorithm, solving large systems of linear equations using the Harrow–Hassidim–Lloyd algorithm, and simulating quantum many-body systems by exploiting superposition for exponential state spaces.In stark contrast, machine learning thrives on generalizing patterns from incomplete, often noisy sample data drawn from complex, real-world distributions, with the goal of making accurate predictions on unseen instances without relying on predefined structures. Due to this fundamental mismatch in their natures—quantum's reliance on precise interference versus machine learning's data-driven adaptability—directly mapping machine learning methods onto quantum computers can feel forced and contrary to the principles of quantum algorithms[13].

A more promising direction is a hybrid approach that integrates quantum computing with modern AI to enhance performance in targeted ways. For instance, Grover's algorithm offers a quadratic speedup for search problems, making it valuable for accelerating exploration in reinforcement learning by efficiently identifying optimal solutions from vast possibilities[39]. Additionally, the expressive capacity of qubits can be leveraged for more precise feature extraction within classical machine learning pipelines [40].

Ultimately, while these hybrid advances show potential, they are unlikely to supplant existing DNNs. Instead, the two paradigms will most likely coexist and complement one another.

## 3. Challenges of SNNs

### 3.1 Limitations of SNNs on Language Tasks

Modern Transformer-based models, which excel in language tasks, employ attention mechanisms to simultaneously consider all token-to-token relationships, thereby maintaining long-range dependencies [14]. In contrast, SNNs rely primarily on discrete spike events and timing, limiting their representational bandwidth. This event-driven paradigm makes it difficult to encode rich semantic relationships or manipulate token interactions. As a result, SNNs struggle to capture long-range dependencies or preserve continuous context—both of which are crucial for language comprehension [15]. While biological brains achieve a "rich effective bandwidth" through a combination of spikes, synchrony, oscillations, neuromodulatory signaling, and dynamic cross-regional interactions [41], current SNN models fail to replicate this complexity due to several key shortcomings, including overlooking neuronal heterogeneity, which hinders emulation of diverse spiking behaviors and dendritic computations, and inadequately incorporating cell-type specific neuromodulatory effects essential for multi-scale learning and adaptability [42].

### 3.2 Limited Value of Faithful Brain Emulation

The human brain, shaped by biological evolution, operates under various constraints that result in redundant and latency-prone cognitive processes. For example, its cell-based metabolic systems rely on finite chemical reactions, primarily using ATP (adenosine triphosphate) for energy. This reliance severely limits overall energy consumption and enforces sparse information representations to avoid excessive neural firing [16].

In its evolutionary context, the brain developed primarily to solve immediate, real-world challenges like survival, predator avoidance, and reproduction. This has left abstract computational abilities in a relatively underdeveloped state. Consequently, human intelligence is burdened by structural limitations, including cognitive biases (e.g., anchoring, confirmation bias), limited working memory, and an inability to multitask effectively. To handle complex tasks, the brain must often resort to indirect approaches with redundant steps, which reduce processing speed and efficiency [19].

For instance, a DNN can train a ResNet-50 model on the ImageNet dataset to 75.3% accuracy in just 74.7 seconds [17], whereas humans require months or years for comparable learning. Similarly, a DNN using fastText can learn word embeddings and analogies from over a billion words in under 10 minutes [18], a feat that takes human children years [16]. Therefore, faithfully emulating these inherent inefficiencies would deliberately impose unnecessary weaknesses on AI and hinder its performance [19].

Lastly, if intelligence is defined as "the capacity to realize complex goals," then it can be understood as taking multifaceted and diverse forms. Human intelligence represents merely one variant, shaped by biological evolution and constrained by its inherent limitations, rather than the ultimate benchmark to be emulated. Therefore, striving to replicate the human brain risks falling into an anthropocentric trap. This approach overlooks the potential to develop AI that capitalizes on silicon-based strengths—such as ultra-fast computation (processing signals near the speed of light, far exceeding human neural conduction at 120 m/s), vast memory capacity (storing and retrieving petabytes of data without decay or forgetting), and seamless scalability (instantly upgrading hardware, reconfiguring algorithms, or copying learned skills across systems without biological constraints)—to forge novel forms of intelligence that surpass human abilities in key domains[19].

### 3.3 Limited Energy Efficiency Advantages of SNNs Compared to Optimized DNNs

Although SNNs are often praised for their potential energy efficiency, recent studies reveal that optimized DNNs using techniques like quantization can be more efficient in practice.

Yan et al. [20] re-evaluated this claim by creating a fair comparison between SNNs and their equivalent Quantized Neural Networks (QNNs). They mapped rate-encoded SNNs with T timesteps to QNNs with $[log_2(T+1)]$ bits. Their energy cost analysis—accounting for computation, memory access, and data movement—found that SNNs are only more efficient under very strict conditions, such as spike rates below 6.4% and short time windows of T=5–10. Otherwise, optimized QNNs consume less energy due to reduced data movement overheads from dense and static computation patterns, as well as more predictable memory access patterns enabling better hardware utilization.

Complementing this, Shen et al.[21] found that while SNNs with multiple time steps are analogous to multi-bit QNNs, the latter often outperform SNNs in low-latency settings. By strategically allocating bits to weights and activations, QNNs can achieve comparable or superior accuracy (e.g., 96.84% on CIFAR-10 with a 4-bit configuration) at a lower computational cost. Together, these findings underscore that the energy efficiency advantages of SNNs are not always realized in practice and can be surpassed by well-optimized QNNs.

### 3.4 High Computational and Memory Overhead in SNN Training

SNNs present significant practical challenges in terms of training overhead. Unlike DNNs, which process inputs through a single forward-backward pass, SNNs require temporal unfolding across numerous time steps. This necessity to update membrane potentials and propagate errors back through time at each step drastically increases both computational complexity and memory requirements[22]. These are not minor implementation hurdles but fundamental issues that even specialized neuromorphic hardware may not fully resolve[23]. Consequently, a common workaround is to first train a conventional DNN and then convert its parameters to an SNN architecture. While this approach leverages the mature DNN training ecosystem to bypass the difficulties of direct SNN training[21], this very dependence highlights the fundamental challenge SNNs face in truly supplanting them.

**4 Future Prospects of DNNs**

**4.1 The Rise of LRMs**

*4.1.1 Mechanisms and Benefits of LRMs*

The prevailing strategy for advancing LLMs was once thought to be simple: follow scaling laws. This approach assumed that continually enlarging model parameters and training datasets would deliver steady performance gains. However, recent analyses have revealed a significant obstacle: improvements are projected to stall as the supply of high-quality, publicly available training data is depleted[27].
The advent of LRMs, however, marks a significant shift in this trajectory. By leveraging extended internal thought processes for complex reasoning, LRMs have driven substantial performance gains and established a new state-of-the-art (SOTA) for the industry, continually advancing benchmarks in fields like mathematics, science, and coding[24]. Most recently, xAI's Grok-4 Heavy, released on July 9, 2025, demonstrated this progress by achieving SOTA results of 88.9% on GPQA, 100% on AIME 2025, and 44.4% on Humanity's Last Exam (with tools) [26]. Notably, unlike traditional LLMs that rely on model size and training compute for scaling, LRMs achieve performance scaling by increasing compute resources at inference time rather than during training[28].
LRMs leverage RL to autonomously generate high-quality reasoning traces. By employing algorithms such as MCTS, they significantly reduce their dependency on expensive human-annotated data. This process creates a self-improving loop: the models iteratively refine their outputs by generating, evaluating, and learning from their own reasoning through RL feedback and search-based exploration. Consequently, LRMs overcome the data limitations of traditional scaling laws and enable a sustainable cycle of performance improvement in both training and evaluation[25].
Furthermore, reasoning models can achieve superior results compared to traditional LLMs under equivalent total compute budgets by optimizing performance through additional inference-time computation on smaller base models[29]. For instance, in FLOPs-matched evaluations on the MATH benchmark, a smaller PaLM 2-S model augmented with test-time compute outperforms a ~14× larger pretrained model, achieving relative accuracy improvements of up to 27.8% on questions of medium difficulty when the inference-to-pretraining token ratio (R) is low (R≪1). This efficiency is particularly significant in scenarios like self-improvement pipelines or on-device deployment, where inference tokens are substantially fewer than pretraining tokens (R≪1). This enables the iterative refinement of model outputs with reduced human supervision and allows smaller models to substitute for larger ones, thereby lowering the environmental and cost burdens associated with extensive pretraining[29].

These benefits align well with advancements in inference-focused ASICs, as discussed in the next section, like those from Groq and Sohu, which amplify the efficiency of compute-intensive reasoning during deployment.

Complementing this, distillation of reasoning models is more effective than that of traditional LLMs. This superiority arises because reasoning distillation focuses on capturing step-by-step thought processes and explanation traces, such as through chain-of-thought methods like CoT-Distill. In contrast to traditional distillation, which primarily replicates final outputs, this focus on cognitive patterns enables student models to generalize better and handle complex, multi-step tasks[30]. Moreover, techniques like quantization and MoE architectures are enabling high-performance reasoning in more compact models. MoE, for example, selectively activates a subset of specialized sub-networks ("experts") for each token to achieve high performance with fewer active parameters. This progress is exemplified by recent releases like the 120B-parameter gpt-oss-120b, which rivals leading industry models in reasoning tasks while remaining deployable on a single 80GB GPU. For instance, it achieved 97.9% on AIME 2025 (with tools), 80.1% on GPQA Diamond (without tools), and 90.0% on MMLU. In these benchmarks, it surpassed OpenAI's o3-mini and approached the performance of o4-mini[31].

4.1.2 *Addressing Criticisms and Skepticism*

Despite this promise, some researchers express skepticism, arguing that LRM performance degrades on highly complex tasks. They cite inefficiencies such as "overthinking," where a model identifies a correct solution but continues to waste resources exploring incorrect alternatives, and "fixation on early errors," where it squanders its computational budget by clinging to a flawed initial hypothesis[43].

However, these appear to be growing pains rather than fundamental limitations, as such issues have steadily been mitigated through targeted advancements. For instance, cutting-edge reinforcement learning approaches, such as Group Relative Policy Optimization (GRPO) implemented in DeepSeekMath 7B, strengthen mathematical reasoning by curbing error accumulation across sequential steps, while test-time compute scaling adaptively optimizes reasoning trajectories to prevent overthinking and improve efficiency[44]. Additionally, innovations in reinforcement learning that use length-based rewards, supervised fine-tuning on variable-length chain-of-thought data, and dynamic inference paradigms that adaptively truncate unnecessary steps have demonstrably reduced overthinking. These methods enable shorter yet effective reasoning sequences in models like DeepSeek-R1 and QwQ-32B without sacrificing performance on benchmarks such as MATH-500 and GSM8K[45]. Consequently, the industry continues to witness state-of-the-art performance from LRMs, as exemplified by recent models like Grok-4 Heavy[26].

A related critique of Shojaee et al.[43] highlights methodological flaws in their evaluations, particularly the failure to account for output token constraints. For example, their automated assessments often misinterpret a model's deliberate truncation of a response—a practical step to avoid exceeding a context window—as a reasoning failure[46]. In their Tower of Hanoi experiment, models failed when token limits (e.g., 64k) were breached, leading to explicit omissions like, "The pattern continues, but to avoid making this too long, I'll stop here." These outcomes were erroneously classified as cognitive breakdowns rather than practical truncations. Similarly, in the River Crossing puzzle, their scoring method inflates the perception of failure by marking instances where N ≥ 6 as errors, even though such scenarios are mathematically impossible to solve[46].

Finally, Shojaee et al.[43] raise questions about whether LRMs are truly capable of generalizable reasoning or if they primarily rely on sophisticated forms of pattern matching, potentially limiting their ability to handle novel problems and challenging their status as intelligent systems. This perspective, however, prompts a deeper examination of intelligence itself. As Mattson[47] argues, human cognition is fundamentally rooted in pattern recognition, where the brain's advanced processing enables creativity, language, and imagination through encoding, integrating, and transferring patterns—mechanisms that closely parallel AI learning via gradient-based training and may even mirror neurobiological mechanisms[47-48]. If dependence on pattern matching disqualifies LRMs from true reasoning, it would similarly undermine our view of human intelligence, which operates on comparable foundations.

In a nutshell, the concerns highlighted by Shojaee et al.[43] more likely stem from current implementation limitations rather than an insurmountable barrier to LRMs' reasoning capabilities.

### 4.2 The Shift to Specialized ASICs for Inference

While GPUs currently dominate the AI landscape, the future points towards a significant shift to ASICs. The maturation of ASICs is poised to enhance energy efficiency, reduce costs, and accelerate computation, thereby propelling AI's evolution. This transition mirrors historical precedents, such as the shift in Bitcoin mining from CPUs and GPUs to ASICs, which improved computational efficiency by thousands of times[36]. A similar boost is expected in AI, particularly because inference tasks involve fixed computational patterns, making them ideal for specialized hardware design[37-38].

Concrete examples already highlight this potential. The Cerebras Wafer Scale Engine (WSE), for instance, reduces inference latency by 10–20 times compared to GPUs while improving energy efficiency 2.5-fold at cost parity[32]. Similarly, Groq's Language Processing Unit (LPU), according to the company's self-reported benchmarks, achieves 300 tokens per second per user on Meta's Llama-2 70B model, enabling low-latency inference that is 10 times the speed of traditional GPU setups while being 10 times more energy efficient[33]. Independent research has confirmed this advantage, showing Groq's system achieves up to 20x lower inference latency than NVIDIA A100 GPUs on models like GPT-2[34]. Emerging solutions such as Etched's Sohu further exemplify these advantages; although based on pre-commercial claims ahead of its 2025 release, a single 8xSohu server is projected to serve over 500,000 Llama 70B tokens per second, matching the performance of 160 H100 GPUs while being an order of magnitude faster and cheaper than NVIDIA's B200[35]. These developments underscore how specialized ASICs can solidify the advantages of DNNs over alternatives like QML or SNNs by optimizing hardware for established neural network paradigms.

## 5. Conclusion

In conclusion, while QML and SNNs have garnered substantial hype for their potential to revolutionize AI through quantum speedups and brain-like energy efficiency, their practical limitations render them unlikely to supplant DNNs in the foreseeable future.

QML faces steep hurdles in adapting backpropagation due to unitary constraints, state collapse, barren plateaus, and measurement overheads, compounded by the limitations of current NISQ hardware, the risk of overfitting due to underdeveloped regularization methods, and a fundamental misalignment with machine learning's data-driven paradigm.

Similarly, SNNs are constrained by limited representational bandwidth for language tasks, the inefficiencies of faithful brain emulation (including cognitive biases and slow learning), overstated energy advantages relative to optimized DNNs, and high training overheads that often necessitate reliance on DNN conversions.

In contrast, DNNs benefit from a mature ecosystem featuring efficient backpropagation, robust regularization techniques, and ongoing innovations in LRMs that mitigate the data scarcity limitations of traditional scaling laws by shifting the focus to inference-time compute, which enables self-improving loops via reinforcement learning and search algorithms. These advancements, coupled with the rise of specialized ASICs like Cerebras WSE, Groq LPU, and Etched Sohu, promise dramatic gains in efficiency, latency, and cost, solidifying DNNs' dominance. Recent benchmarks from models like gpt-oss-120b and Grok-4 Heavy further underscore this trajectory, demonstrating that DNNs can achieve state-of-the-art performance without the exotic hardware or paradigm shifts required by QML or SNNs.

Ultimately, QML and SNNs may find niche roles in hybrid systems, complementing rather than replacing DNNs. The path forward lies in leveraging DNNs' proven strengths while addressing sustainability through continued optimization, ensuring AI's evolution remains grounded in practicality over speculation. This analysis, drawing on 2025's latest research and industry trends, highlights the need for tempered enthusiasm and a focus on scalable, deployable solutions to drive meaningful progress in AI.

**Data availability**

No data are associated with this article

**Competing interests**

No competing interests were disclosed.

**Grant information**

The author(s) declared that no grants were involved in supporting this work.

**Acknowledgements**

The author has no acknowledgments to declare.